# CMOS-Free Multilayer Perceptron Enabled by Four-Terminal MTJ Device

Wesley H. Brigner[1], Naimul Hassan[1], Xuan Hu[1], Christopher H. Bennett[2], Felipe Garcia-Sanchez[3,4], Matthew J. Marinella[2], Jean Anne C. Incorvia[5], Joseph S. Friedman[1]
[1]Electrical and Computer Engineering, University of Texas at Dallas, Richardson, TX, United States
[2]Sandia National Laboratories, Albuquerque, NM, United States
[3]Department of Applied Physics, University of Salamanca, Salamanca, Salamanca, Spain
[4]Istituto Nazionale di Ricerca Metrologica, Turin, Piedmont, Italy
[5]Electrical and Computer Engineering, University of Texas at Austin, Austin, TX, United States

*Abstract*—Neuromorphic computing promises revolutionary improvements over conventional systems for applications that process unstructured information. To fully realize this potential, neuromorphic systems should exploit the biomimetic behavior of emerging nanodevices. In particular, exceptional opportunities are provided by the non-volatility and analog capabilities of spintronic devices. While spintronic devices have previously been proposed that emulate neurons and synapses, complementary metal-oxide-semiconductor (CMOS) devices are required to implement multilayer spintronic perceptron crossbars. This work therefore proposes a new spintronic neuron that enables purely spintronic multilayer perceptrons, eliminating the need for CMOS circuitry and simplifying fabrication.

## I. Introduction

In the human brain, a neuron integrates a series of electrical spikes through its axons and, when enough of these current pulses have been integrated, it releases a spike of its own from the soma (cell body) and through its dendrites to the axons of other neurons. Synapses, on the other hand, provide electrical connectivity between the axons of one neuron and the dendrites of other neurons.

To mimic the brain, neuromorphic systems typically contain artificial neuron and synapse analogs. Such systems can be emulated using software on von Neumann computers [1],[2]; however, due to the fact that complementary metal oxide semiconductor (CMOS) technology, which is the primary technology used in standard von Neumann computers, was not specifically designed to implement the required behavior, these systems consume considerably more energy than an actual human brain [3]. Novel CMOS-based systems can also be used to implement neuromorphic systems [4],[5], but again, CMOS was not specifically designed to match the needs for neuromorphic computing and such chips are highly inefficient.

To resolve the issue of inefficiency and high power consumption, neuromorphic-inspired beyond-CMOS devices have garnered a considerable amount of attention in the scientific community. Since synapses only require non-volatility and a variable resistance, several beyond-CMOS synapses have already been proposed [6]-[9]. On the other hand, fewer beyond-CMOS neurons have been proposed due to the unique challenges in emulating the leaky integrate-and-fire (LIF) neuron model [10]-[12]. While these recently proposed spintronic neurons efficiently provide leaking, integrating, firing, and lateral inhibition, electrical connectivity between the input and output ports results in a need for CMOS circuitry between perceptron layers.

This paper therefore proposes a novel spintronic LIF neuron that enables a monolithic, purely spintronic neuromorphic architecture that leverages previously proposed spintronic synapses. A brief background into the field of neuromorphic computing is provided in section II, and background on spintronic synapses and our previously proposed neurons is presented in section III. Section IV discusses the new spintronic neuron, section V proposes the CMOS-free spintronic multilayer perceptron, and conclusions are provided in section VI.

## II. Background

Due to the fact that novel technologies and architectures are still required to be compatible with currently existing fabrication techniques, many beyond-CMOS technologies, including the neuromorphic structure proposed in this work, are designed using a crossbar array structure.

### A. Crossbar Array

An NxM crossbar array consists of N horizontal wires (word lines) and M vertical wires (bit lines). For a layer in a neuromorphic system, a single input neuron will be attached to each word line and a single output neuron will be attached to each bit line, resulting in N+M total neurons. Likewise, a single synapse exists at each word and bit line intersection to provide connectivity between the input and output neurons for a total of N*M synapses. The resistance states of these synapses, which are capable of being finely tuned during training, determine the weighting between the input and output neurons and, ultimately, the overall functionality of the neuromorphic layer [13]-[15].

### B. Leaky Integrate-and-Fire Neuron

A biologically-accurate neuron model is the leaky integrate-and-fire (LIF) neuron model, which is a considerable improvement over the previous integrate-and-fire (IF) neuron [1]. LIF neurons should implement three primary functionalities. As implied by the name, these functionalities are integrating, leaking, and firing. When an energy spike is fed into the input of an LIF neuron, the neuron should store the energy from the spike. However, when no energy is applied to the input of the neuron, it should gradually dissipate the stored energy. Finally, once the stored energy has reached a certain threshold, the neuron should suddenly release an output energy spike and reset its state to the initial state.

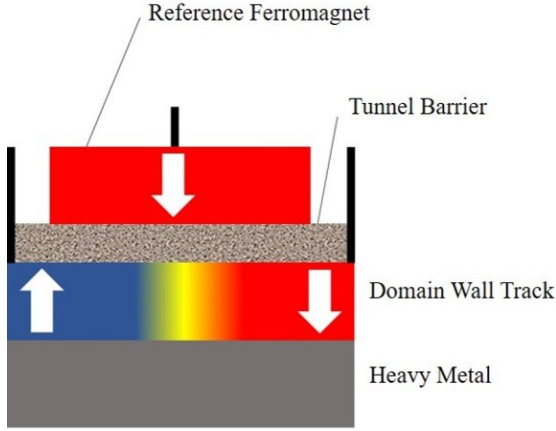

Fig. 1. DW-MTJ synapse with wide tunnel barrier for analog resistance states.

However, integrating, leaking, and firing are not the only functions an LIF neuron should be able to implement. One such secondary functionality is lateral inhibition, where one neuron that has integrated more than nearby neurons should inhibit integration by its neighbors. This functionality can be extended to implement a winner take all (WTA) system, where a single neuron firing causes a reset of all other neurons, allowing only one neuron to fire for a given data set [10].

### III. DOMAIN WALL MAGNETIC TUNNEL JUNCTION

A domain wall magnetic tunnel junction (DW-MTJ) consists of a "free" ferromagnetic layer containing two magnetic domains separated by a domain wall (DW) such that the position of the DW wall determines the MTJ resistance. When the DW, which can be moved using a current, passes underneath, the MTJ switches between the parallel (conductive) and anti-parallel (resistive) states.

#### A. DW-MTJ Synapse & Neuron Integration

This synapse consists of a long tunnel barrier that covers a large portion of the DW track [8], [9], allowing the device to exhibit analog resistance states (Fig. 1). When a current flows through the DW track, the DW shifts from one end of the track to the other. When used in a DW-MTJ neuron, this DW motion represents integration through an increase in stored energy.

#### B. Leaking with a DW-MTJ Neuron

Leaking can be achieved using one of three methods. With the first method, a ferromagnet is placed underneath the DW track, as illustrated in Figure 2a. When an external magnetic field is applied parallel to a magnetic domain's magnetization state, the field causes the domain to expand. Conversely, if the same magnetic field is applied antiparallel to a magnetic domain's magnetization state, the field will cause the domain to shrink. Therefore, a magnetic field applied to a DW track along the same axis as the two antiparallel magnetic domains will cause the DW to shift from one side of the track to the other. The dipolar coupling field provided by the aforementioned ferromagnet allows the DW to shift from one end of the track to the other in the absence of any current. As illustrated in the micromagnetic simulation results of Fig. 3a, when a current is applied from right to left through the DW track, the DW shifts from left to right. However, when no current is applied, the DW gradually shifts from right to left [10].

In the second method, the DW track contains a magnetocrystalline anisotropy gradient, which can be implemented using a thickness and/or composition gradient [16]-[18]. When the DW is in a region of the track with a higher anisotropy value, it exists in a higher energy state, and when the DW is in a region of the track with a lower anisotropy value, it exists in a lower energy state. Therefore, the anisotropy gradient generates an energy landscape that is more favorable to the DW existing on one side of the track than the other. Fig. 3b demonstrates micromagnetic simulation results of the combined leaking and integrating characteristics of a graded anisotropy neuron. As with the neuron with dipolar coupling, a current passed from right to left in the device causes a rapid shift in the DW position from left to right. When no current is applied, the DW slowly leaks from right to left [11].

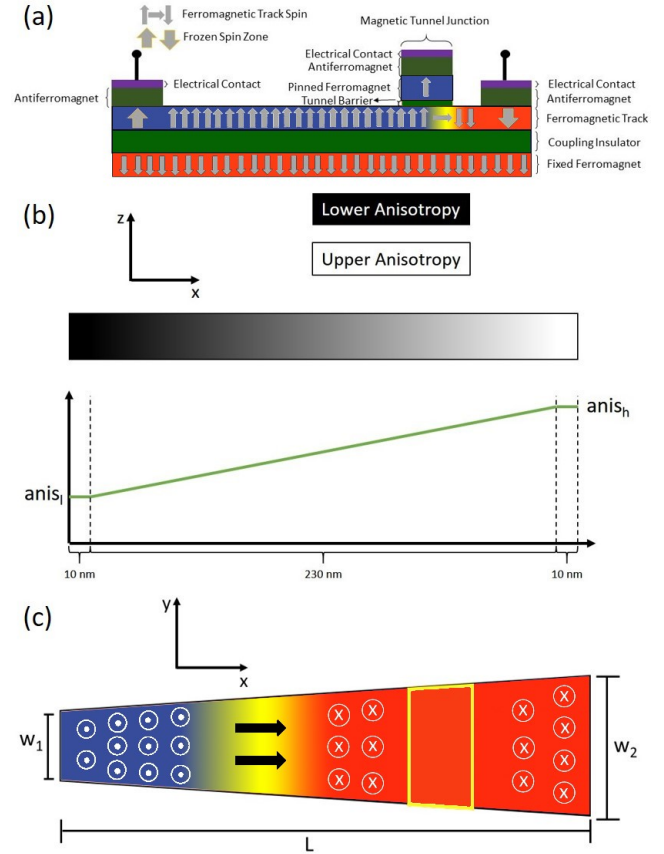

Fig. 2. (a) Side view of the neuron with dipolar coupling field. (b) Side view of the neuron with graded anisotropy illustrating the anisotropy gradient instead of magnetization. (c) Structure of the neuron with shape-based DW drift.

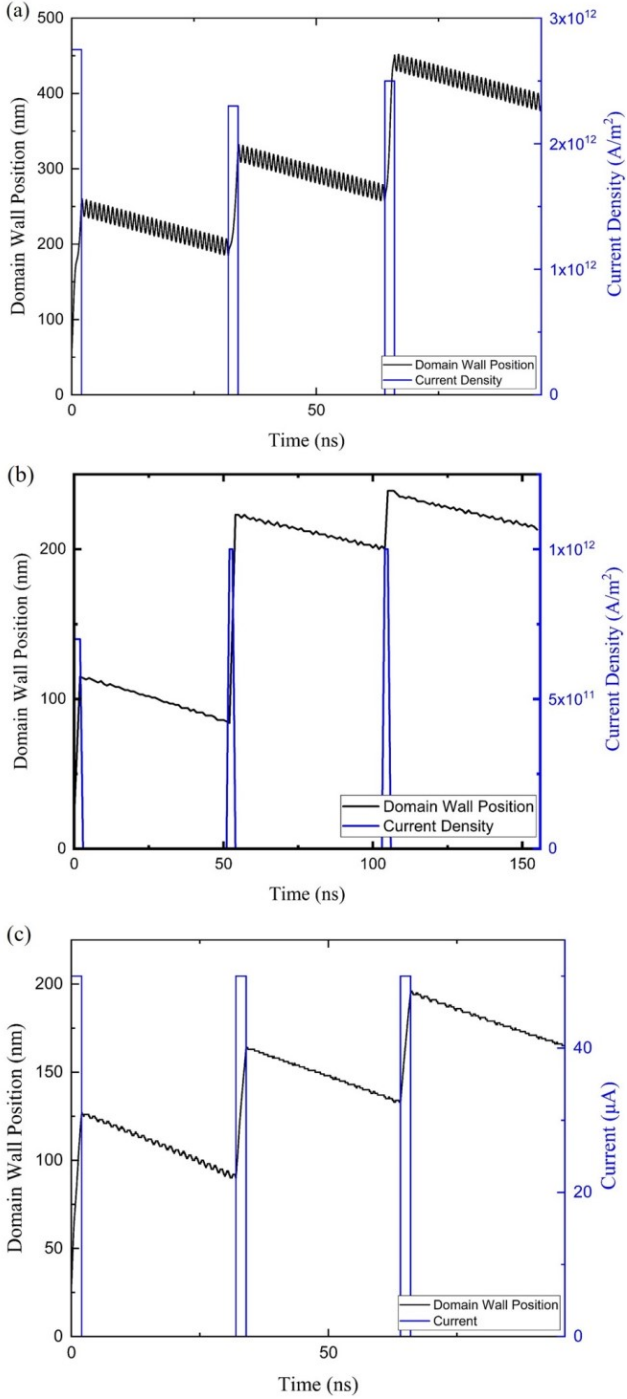

Fig. 3. (a) Combined integrating and leaking characteristics of a neuron implemented using a dipolar coupling field. (b) Combined integrating and leaking characteristics of a neuron with anisotropy gradient. (c) Combined integrating and leaking characteristics of a neuron with shape-based DW drift.

Finally, with the third method, the sides of the DW track are angled to form a trapezoidal track instead of a rectangular track. Similar to the anisotropy gradient neuron, the DW exists in a higher energy state when it is in the wider region than when it is in the narrower region. Therefore, the DW will shift from the wide portion of the DW track to the narrow portion without any externally-applied stimuli. When applying a current from right to left through the device, as in Fig. 3c, the DW shifts from left to right. When no current is passed through the device, the energy landscape present in the DW track forces the DW to shift from right to left [12].

## IV. Firing with Four-Terminal DW-MTJ Neuron

In order to avoid electrical connectivity between the input and output ports of the DW-MTJ neuron, an addition free ferromagnetic layer is added in the novel four-terminal DW-MTJ LIF neuron proposed in Fig. 4. When the DW passes underneath the electrically-isolated MTJ, the free layer of the MTJ reorients itself to match the magnetization of the DW track underneath it due to dipolar coupling, thereby switching the state of the MTJ. This effect, in combination with a voltage applied at the top terminal, provides the firing functionality without any electrical connectivity between the MTJ and DW track.

## V. CMOS-Free Multilayer Spintronic Perceptron

This electrical isolation within the four-terminal DW-MTJ neuron enables the first CMOS-free spintronic multilayer perceptrons, as shown in Fig. 5. While spintronic neural networks have previously been considered [9], significant CMOS circuitry was required in addition to the spintronic synapses and neurons in order to interconnect the layers, implement the leaking, and provide lateral inhibition.

As shown in Fig. 5, the electrical isolation provided by the four-terminal DW-MTJ allows an M×N crossbar layer to be connected to an N×O crossbar layer while maintaining unidirectional signal flow. This CMOS-free spintronic structure can be extended to deep neural networks with numerous layers.

## VI. Conclusions

This work proposes three novel DW-MTJ neurons that, in conjunction with DW-MTJ-based synapses, enable a purely spintronic neural network. This LIF neuron intrinsically provides the leaking, integrating, and firing capabilities, thereby eschewing the need for additional CMOS circuitry.

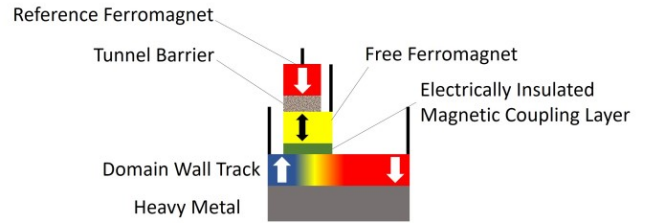

Fig. 4. Schematic of the four-terminal DW-MTJ LIF neuron. When a current flowing through the DW track shifts the DW underneath the MTJ, the magnetization of the free ferromagnet switches, thereby switching the MTJ resistance state. With a constant supply voltage applied across the MTJ, this state change results in increased current output through the MTJ.

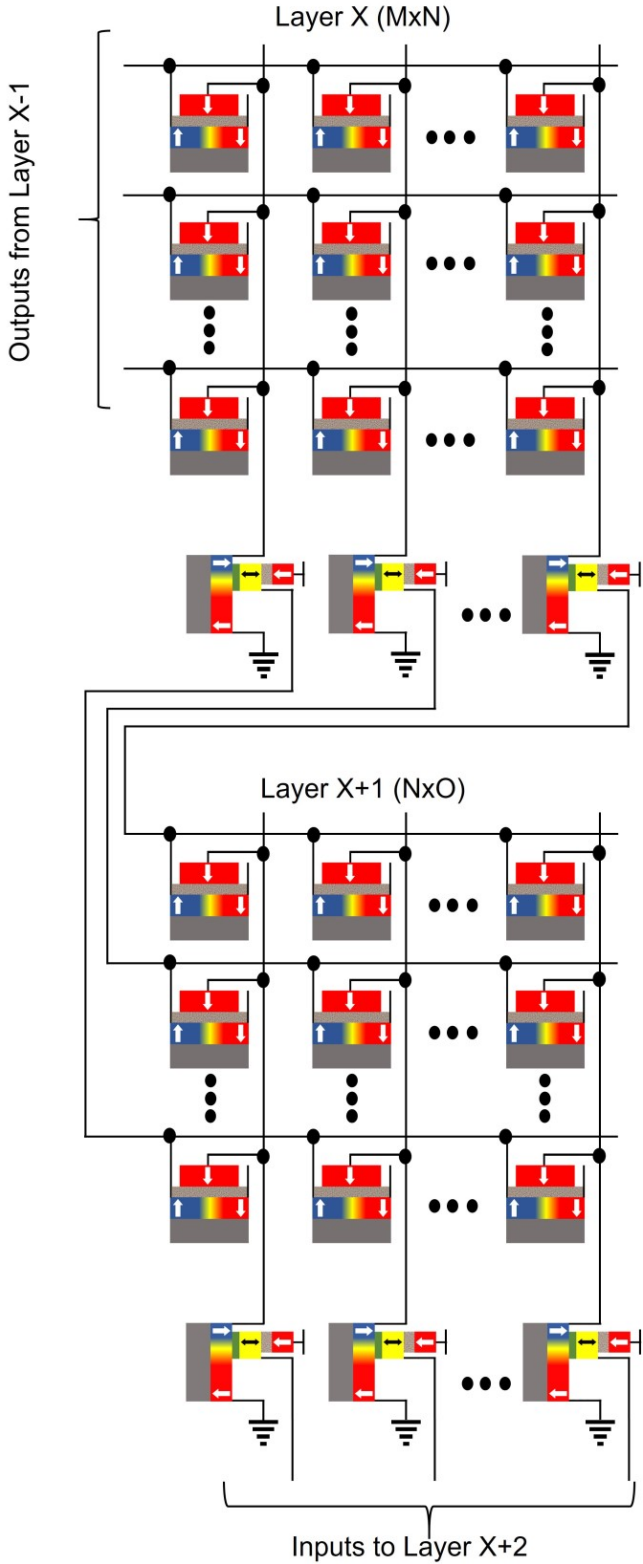

Fig. 5. Multilayer perceptron using the four-terminal DW-MTJ neuron of Fig. 4 and the three-terminal DW-MTJ synapse of Fig. 1. The synapses at the intersections of word and bit lines provide weights between two individual neurons, and the neurons of one layer provide the inputs to another layer.

Therefore, we have the capability of implementing monolithic CMOS-free multilayer spintronic perceptrons. By exploiting biomimetic synapses and neurons in a simplified fabrication, this multilayer perceptron concept promises significant advances for efficient machine learning and artificial intelligence.

DISTRIBUTION STATEMENT

DISTRIBUTION STATEMENT A. Approved for public release: distribution is unlimited.

ACKNOWLEDGMENTS

This research is sponsored in part by the National Science Foundation under CCF awards 1910800 and 1910997. The authors thank E. Laws, J. Mcconnell, N. Nazir, L. Philoon, and C. Simmons for technical support, and the Texas Advanced Computing Center at The University of Texas at Austin for providing computational resources. Sandia National Laboratories is a multimission laboratory managed and operated by NTESS, LLC, a wholly owned subsidiary of Honeywell International Inc., for the U.S. Department of Energy's National Nuclear Security Administration under contract DE-NA0003525. This paper describes objective technical results and analysis. Any subjective views or opinions that might be expressed in the paper do not necessarily represent the views of the U.S. Department of Energy or the United States Government.